\documentclass[10pt,conference,compsocconf]{IEEEtran}
\usepackage{times}  % DO NOT CHANGE THIS
\usepackage{helvet} % DO NOT CHANGE THIS
\usepackage{courier}  % DO NOT CHANGE THIS
\usepackage[hyphens]{url}  % DO NOT CHANGE THIS
\usepackage[ruled, vlined, algo2e]{algorithm2e}
\usepackage{graphicx} % DO NOT CHANGE THIS
\usepackage{smile}
\usepackage{graphicx}  % DO NOT CHANGE THIS
\newcommand{\resnet}{\emph{ResNet-18}\xspace}
\newcommand{\mobilenet}{\emph{MoibleNet}\xspace}
\newcommand{\densenet}{\emph{DenseNet}\xspace}
\newcommand{\preact}{\emph{PreActResNet-18}\xspace}
\newcommand{\googlenet}{\emph{GoogLeNet}\xspace}
\newcommand{\dpn}{\emph{DPN}\xspace}
\newcommand{\senet}{\emph{SENet}\xspace}
\newcommand{\kavg}{KAVG\xspace}
\newcommand{\mavg}{MAVG\xspace}
\newcommand{\cifar}{\emph{CIFAR-10}\xspace}
\newcommand{\imagenet}{\emph{ImageNet-1K}\xspace}

\title{Accelerate Distributed Stochastic Descent for Nonconvex Optimization with Momentum }
\author{
   \IEEEauthorblockN{Guojing Cong}
  \IEEEauthorblockA{
  IBM TJ Watson Research Center\\
  gcong@us.ibm.com}

\and

\IEEEauthorblockN{Tianyi Liu}
  \IEEEauthorblockA{
  Georgia Institute of Technology\\
  tyliu@gatech.edu}
  }

\begin{document}

\maketitle

\begin{abstract}
Momentum method has been used extensively in optimizers for deep learning. Recent studies show that distributed training through K-step averaging has many nice properties. We propose a momentum method for such model averaging approaches. At each individual learner level traditional stochastic gradient is applied. At the meta-level (global learner level), one momentum term is applied and we call it block momentum. We analyze the convergence and scaling properties of such momentum methods. Our experimental results show that block momentum not only accelerates training, but also achieves better results.

\end{abstract}
\section{Introduction}
Deep neural networks (DNNs) have achieved great success recently in many different domains.  Training DNNs requires solving large-scale non-convex optimization problems. This process is usually time consuming because of the fast increasing size of datasets and complexity of the network structures.   As a result, researchers often wait for weeks or months for models to train. For example, finishing a 90-epoch ImageNet-1k ($1$ million training images in resolution $224 \times 224$) training with large scale ResNet (around $25.6$ million parameters) on a single NVIDIA Titan XP GPU takes over 10 days. To accelerate the training process,  parallel and distributed processing is widely adopted in deep learning applications.

A line of parallel solvers based on Stochastic Gradient Descent \cite{robbins1951stochastic}, or SGD for short, are proposed these years. \cite{zinkevich2010parallelized} first introduces a parallel variants of SGD called synchronous SGD.  Theoretical analysis has shown that synchronous  SGD with $P$ learners achieves a convergence rate of $O(1/\sqrt{NP})$ for solving non-convex optimization problems, where $N$ is the number of samples processed. In 2012, \cite{dean2012large} presents the Asynchronous SGD (Async-SGD). Different from its synchronous counterpart, each worker communicates with the parameter servers independently of the others when running Async-SGD, and thus, the communication cost is reduced. \cite{lian2015asynchronous} shows that Async-SGD (e.g., ~\cite{recht2011hogwild,KP15}) converges for non-convex objectives and achieves the same convergence rate when the staleness is bounded. In practice, however, Async-SGD is hard to implement because of the requirement of small learning rates and the difficulty in controlling the staleness. Recently, \cite{zhou2017convergence} introduces the K-step average SGD (K-AVG) which allows delayed gradient aggregation to minimize the communication overhead. In this algorithm, the communication interval $K$ is manually chosen.  \cite{zhou2017convergence} theoretically and empirically verifies that the optimal $K\neq 1$ for many different cases. A large $K$ greatly reduces the communication cost.   Thus, the negative impact on convergence caused by asynchrony  can be avoided. Moreover, K-AVG scales better and allows larger learning rates than Async-SGD.

Many machine learning models, however, are often trained using algorithms equipped with momentum such as Momentum SGD \cite{polyak1964some} where a weighted average of the  previous update is added to the current gradient. For example, MSGD can achieve the best test accuracy for training ResNet for image classification tasks \cite{HZR16a}.  \cite{ghadimi2016accelerated} show that MSGD is guaranteed to converge to the  first order stationary point for non-convex objective. \cite{yang2016unified} provides a uniform framework to analyze the convergence of SGD with momentum for both convex and non-convex objectives. They further show that MSGD can achieve $O(1/\sqrt{N})$ convergence rate for non-convex objectives. Another line of research tends to study the reason behind the good performance of MSGD. \cite{wilson2016lyapunov} suggest that momentum accelerates computation by avoiding ``long ravines" and ``sharp curvatures" in the sub-level sets of cost function. 
\cite{liu2018toward} theoretically proves that MSGD can help avoid saddle points for the streaming PCA problem.  \cite{ochs2015ipiasco} presents an illustrative example to show that the momentum can potentially avoid local minima. 

Recent years, people start to focus on the application of momentum in parallel computing both empirically and theoretically. Empirical study such as \cite{chen2016scalable,lin2018don}  observe that momentum can improve the convergence and text accuracy for distributed training of neural networks. \cite{yu2019linear} considers a distributed communication efficient MSGD method and proving its linear speedup property. \cite{liu2018towards} shows that there is an acceleration trade-off between momentum and delay while using Async-MSGD.

Based on the previous study, in this paper, we propose the Momentum Average SGD (M-AVG) algorithm to incorporate the advantage of momentum in K-AVG. Specifically, at the meta-level (global learner level), one momentum term is applied to add the weighted average of the previous update to the current one, while at each individual learner level, K-step SGD is implemented.We theoretically study the convergence property of M-AVG algorithm and compare it with K-AVG to show the positive impact of the momentum. The main advantage of adding momentum is the speed-up, i.e., with properly chosen step size, M-AVG can achieve same accuracy using less samples than K-AVG. At the same time, adding momentum will inherit the advantage of K-AVG in reducing optimization cost. In fact, our theoretical analysis and experimental result verify that the optimal $K$  is not equal to 1 even in the presence of the momentum parameter. Moreover, to ease the tuning process in practice, several tuning guidelines are included in the paper.

Our main contributions are summarized as follows:
\begin{itemize}
\item We propose a new momentum method M-AVG and prove the convergence rate.
\item We show that M-AVG can achieve faster convergence rate than K-AVG.
\item Several useful tuning guidelines are provided to ease the implementation.
\end{itemize}

{\it Using an image recognition benchmark, we demonstrate
the nice convergence properties of K-AVG in comparison to
two popular ASGD implementations: Downpour [Dean et
al., 2012] and EAMSGD [Zhang et al., 2015]. In EAMSGD,
global gradient aggregation among learners simulates an elastic force that links the parameters they compute with a center
variable stored by the parameter server. In both Downpour
and EAMSGD, updates to the central parameter server can
also have a K-step delay. On our target platform, when K is
small, K-AVG significantly reduces the communication time
in comparison to Downpour and EAMSGD while achieving
similar training and test accuracies. The training time reduction is up to 50%. When K is large, K-AVG achieves
much better training and test accuracies than Downpour and
EAMSGD after the same amount of data samples are processed. For example, with 128 GPUs, K-AVG is up to about
7 and 2-6 times faster than Downpour and EAMSGD respectively, and achieves significantly better accuracy}

The rest of the paper is organized as follows: In section 2,
we introduce the M-AVG algorithm and the optimization problem we considered. Assumptions needed to analyze SGD methods is inclueded; In section 3, we formally the standard convergence results of M=AVG with fixed stepsize for non-convex objectives. Based
on the convergence result, we analyze the speed up of M-AVG over K-AVG and investigate the optimal choice of K. Some useful tunning guidelines is provided in this section; In section 4, we present our experimental results to validate our analysis.

Notations: Given a vector $ v = (v^{(1)}, \ldots, v^{(d)})^\top\in\RR^{d}$, we define the vector norm: $\norm{v}^2 = \sum_j(v^{(j)})^2$. $P$ denote s number of processors. $K$ denotes the length of the delay. $B_n$ or $B$ denotes the size of mini-batch. $\eta_n$ or $\eta$ denotes the step size.  $\mu$ denotes the momentum parameter. $\xi_{k,s}^j$ denotes the i.i.d. realizations of  a random variable $\xi$ generated by the algorithm on different processors and in different iterations, where $j=1,...,N$, $k=1,...,K,$ and $s=1,..,B.$ 

\section{Algorithm and Model}
We propose to solve the following general non-convex minimization problems:
\begin{align}\label{opt}
\min_{\bw\in\cX}F(w),
\end{align}
where the objective  $F:\RR^m\rightarrow R$ is a continuously differentiable non-convex function over $\cX\in \RR^m.$  Training deep neural networks is a special case of \eqref{opt}, where $F(\bw)=\frac{1}{n}\sum_{i=1}^n \ell (y_i,f(x_i,\bw))$ given $n$ observations denoted by $\{(x_i, y_i)\}_{i=1}^{n}$, where $x_i$ is the $i$-th input feature, $y_i$ is the response and $\ell$ is the loss function.
 We will impose the following regularity conditions on the objective $F,$ throughout our analysis.
\begin{assumption}\label{assumption}

\begin{itemize}
\item $\nabla F$ is $L$-Lipschitz, i.e.,
$$\norm{\nabla F(x)-\nabla F(y)}_2\leq L\norm{x-y}_2,~~\forall x,y.$$
\item $F$ is lower bounded by a constant $F^*.$
\item  Bounded Gradient: $\nabla F$ is bounded, i.e., there exists a constant $M$ such that 
$$\norm{\nabla F(x)}_2^2\leq M, ~~\forall x.$$
\item $F$ is bounded below by a constant $F^*.$
\item Unbiased Estimate: For any fixed parameter $\bw$, the stochastic gradient $\nabla F(\bw; \xi)$ is an unbiased estimator of the true gradient 
	corresponding to the parameter $\bw$, namely, 
	$$
	\EE_{\xi} \nabla F(\bw; \xi) = \nabla F(\bw).
	$$
	\item Bounded Variance: There exist a constant $\sigma \geq 0$ such that,
	$$
	\EE_{\xi} \big\| \nabla F(\bw;\xi)\big\|_2^2 - \big\|\EE_{\xi} \nabla F(\bw;\xi) \big\|_2^2 \leq \sigma^2.
	$$

\end{itemize}
\end{assumption}
Same assumptions have been used in \cite{zhou2017convergence}. To solve problem \eqref{opt}, we propose the following  M-AVG algorithm in Algorithm \ref{alg:avg} based on K-AVG algorithm. Specifically,  Each learner  runs mini-batch SGD for k steps and return the $k-th$ iteration to the server. Then at the meta level, th global learner average these returns,  update the momentum and the weight estimation. Note that in Algorithm \ref{alg:avg}, $\bd$ is actually the product of K-step average gradient and the step size. If we denote  $G_n=\frac{1}{B_n P}\sum_{j=1}^P\sum_{t=0}^{K-1}\sum_{s=1}^{B_n} \nabla F(\bw_{n+t}^j,\xi_{t,s}^j).$ Then the update of M-AVG algorithm can be written as
\begin{align}\label{alg}
\bv_{n+1}&=\mu \bv_n-\eta_n G_n,\\\nonumber
\tilde{\bw}_{n+1}&=\tilde{\bw}_{n}+v_{n+1}
\end{align}
where $\mu$ is the momentum parameter and $\eta_n$ is the step size. We will use the update \eqref{alg} to carry on our theoretical analysis. We further note that since the momentum is added to the meta level, we call this momentum block momentum. We can also add momentum to the learner level. %Please see more discussion in Section \ref{conclusion}. 
  
\begin{algorithm}
	%	\KwData{this text}
	%	\KwResult{Output $\widetilde{\bw}$ }
	initialize $\widetilde{\bw}_1$;\\
	$v\leftarrow0;$\\
	\For{$n=1,...,N$}{
		Processor $P_j$, $j=1,\dots,P$ do concurrently:\\
		set $\bw_n^j=\widetilde{\bw}_n$ ;\\
		\For{$k=1,...,K$}{
			randomly sample a mini-batch of size $B_n$ and update:
			$$
			\bw_{n+k}^j = \bw_{n+k-1}^j - \frac{\gamma_n}{B_n} \sum\limits_{s=1}^{B_n} \nabla F(\bw_{n+k-1}^j;\xi_{k,s}^j)
			$$
		}
		$\ba\leftarrow\frac{1}{P}\sum_{j=1}^P \bw_{n+K}^j$;\\
		$\bd\leftarrow\ba-\widetilde{\bw}_{n} $;\\
		$\bv \leftarrow \mu \bv +d$;\\
	         $\widetilde{\bw}_{n+1} = \widetilde{\bw}_{n}+\bv$;\\
	}
	\caption{M-AVG Algorithm}
\label{alg:avg}
\end{algorithm}

\section{Main Results}
\subsection{Convergence of M-AVG}
We study the convergence of M-AVG algorithm. In Theorem \ref{thm}, we show that the algorithm can converge to first order stationary point. We present the upper bound on the expected average squared gradient norms.

\begin{theorem}\label{thm}
Under Assumption \ref{assumption}, suppose Algorithm \ref{alg:avg} is run with fixed step size $\eta>0,$ batch size $B>0$  and momentum parameter $\mu\in[0,1)$ such that the following condition holds.
 $$1\geq \frac{L^2\eta^2(K+1)(K-2)}{2(1-\mu)^2}+\frac{2\eta LK}{1-\mu}$$
 and 
 $$1-\delta\geq L^2\eta^2/(1-\mu)^2,$$
 for some constant $\delta\in(0,1).$ Then the expected average squared gradient norms of $F$ satisfy the following bounds for all $N\in \NN:$
 \begin{align}\label{ineq_thm}
\frac{1}{N} \sum_{i=1}^{N}\EE\big\| \nabla& F(\widetilde{\bw}_{i})\big\|_2^2\leq \frac{2(1-\mu)(F(w_{1})-F^*)}{N(K-1+\delta)\eta}\notag\\&+\frac{L^2\eta^2\sigma^2(2K-1)K(K-1)}{6(K-1+\delta)B(1-\mu)^2}\notag\\
&+\frac{2LK^2\sigma^2 \eta}{PB(K-1+\delta)(1-\mu)}\left(1+\frac{\mu^2}{2(1-\mu)^2}\right)\notag\\&+\frac{L\eta\mu^2K^2M}{(K-1+\delta)(1-\mu)^3}.
\end{align}
\end{theorem}
\begin{remark}
Note that when $\mu=0$, M-AVG is reduced to $K-AVG$. Our Theorem \ref{thm} is also a generalization of  Theorem 3.1 in \cite{zhou2017convergence}. In fact, take $\mu=0$ in \ref{ineq_thm}, the right hand side is equivalent to that of (3.2)  in \cite{zhou2017convergence}.
\end{remark}
\begin{proof}
	The proof follows the proof idea in \cite{yang2016unified}. Since M-AVG shares the same learner level procedure with K-AVG, some important results in \cite{zhou2017convergence} will also be used in our proof.
	
Define an auxiliary sequence  $z_{n}=\tilde{w}_{n}+\frac{\mu}{1-\mu} v_{n}.$ Then $z$ is updated as follows:
\begin{align}\label{z_update} 
z_{n+1}=z_n-\frac{\eta_n}{1-\mu} G_n.
\end{align}Recall that $\nabla F$ is $L$-Lipschitz, thus we have for any $x, y,$ the following inequality holds.
\begin{align}\label{Lip_ineq}
F(y)\leq F(x)+\nabla F(x)^\top(y-x)+\frac{L}{2}\norm{y-x}_2^2.
\end{align}
Combine \eqref{z_update} and \eqref{Lip_ineq} together, we get
\begin{align}\label{bound1}
&F(z_{n+1})-F(z_n)\notag\\\leq& \nabla F(z_n)^\top(z_{n+1}-z_{n})+\frac{L}{2}\frac{\eta_n^2}{(1-\mu)^2}\norm{G_n}_2^2\notag\\
\leq& -\frac{\eta_n}{1-\mu}\nabla F(z_n)^\top G_n+\frac{L}{2}\frac{\eta_n^2}{(1-\mu)^2}\norm{G_n}_2^2\notag\\
=&-\frac{\eta_n}{1-\mu}\left(\nabla F(z_n)-\nabla F(\tilde{\bw}_n)\right)^\top G_n\notag\\&-\frac{\eta_n}{1-\mu}\nabla F(\tilde{\bw}_n)^\top G_n+\frac{L}{2}\frac{\eta_n^2}{(1-\mu)^2}\norm{G_n}_2^2\notag\\
\leq& \frac{1}{2L} \norm{\nabla F(z_n)-\nabla F(\tilde{\bw}_n)}_2^2+\frac{L}{2}\frac{\eta_n^2}{(1-\mu)^2}\norm{G_n}_2^2\notag\\&-\frac{\eta_n}{1-\mu}\nabla F(\tilde{\bw}_n)^\top G_n+\frac{L}{2}\frac{\eta_n^2}{(1-\mu)^2}\norm{G_n}_2^2\notag\\
=&\frac{1}{2L} \norm{\nabla F(z_n)-\nabla F(\tilde{\bw}_n)}_2^2-\frac{\eta_n}{1-\mu}\nabla F(\tilde{\bw}_n)^\top G_n\notag\\&~~~~+\frac{L\eta_n^2}{(1-\mu)^2}\norm{G_n}_2^2.
\end{align}
We only need  bound the three terms on the left hand side of \eqref{bound1}. We start from the first term. By the Lipschitz continuity of $\nabla F,$ we have 
\begin{align*}
\frac{1}{2L}\norm{\nabla F(z_n)-\nabla F(w_n)}_2^2&\leq \frac{L}{2}\norm{z_n-w_n}_2^2\\&=\frac{L}{2}\frac{\mu^2}{(1-\mu)^2}\norm{v_n}_2^2.
\end{align*}
Thus, we only need bound the norm of $v_n.$ Denote $\Gamma_{n-1}=\sum_{i=0}^{n-1}{\mu^i }=\frac{1-\mu^n}{1-\mu}\leq\frac{1}{1-\mu}.$ We can rewrite $v_n$ as the weighted sum  of $\{G_i\}_{i=0}^{n-1}.$ Specifically, we have
\begin{align*}
v_n=\mu v_{n-1}-\eta_{n-1} G_{n-1}=\sum_{i=0}^{n-1}{\mu^i \eta_{n-i} G_{n-i}}.
\end{align*}
Then, the expected norm of $v_n$ can be bounded as follows.
\begin{align*}
\EE \norm{v_n}_2^2&=\EE \norm{\sum_{i=0}^{n-1}{\mu^i \eta_{n-i} G_{n-i}}}_2^2\\&=\Gamma_{n-1}^2\EE \norm{\sum_{i=0}^{n-1}{\frac{\mu^i}{\Gamma_{n-1}} \eta_{n-i} G_{n-i}}}_2^2\notag
\\&\leq\Gamma_{n-1}^2{\sum_{i=0}^{n-1}{\frac{\mu^i}{\Gamma_{n-1}} \eta_{n-i} ^2\EE \norm{G_{n-i}}_2^2}}\\&=\Gamma_{n-1}{\sum_{i=0}^{n-1}{{\mu^i} \eta_{n-i} ^2\EE\norm{G_{n-i}}_2^2}}.
\end{align*}
By the bounded gradient and variance assumption, we have 
\begin{align}\label{normbound}
\EE \norm{G_{n-i}}_2^2&=\EE \norm{\frac{1}{B P}\sum_{j=1}^P\sum_{t=0}^{K-1}\sum_{s=1}^{B} \nabla F(\bw_{n-i+t}^j,\xi_{t,s}^j)}_2^2\notag\\
&\leq \frac{K}{P^2B^2}\sum_{t=0}^{K-1}\EE \norm{\sum_{j=1}^P\sum_{s=1}^{B} \nabla F(\bw_{n-i+t}^j,\xi_{t,s}^j)}_2^2\notag\\
&\leq  \frac{K}{P^2B^2}\sum_{t=0}^{K-1}\EE \norm{\sum_{j=1}^P\sum_{s=1}^{B} \Big(\nabla F(\bw_{n-i+t}^j,\xi_{t,s}^j)\notag\\
	&-\nabla F(\bw_{n-i+t}^j)+\nabla F(\bw_{n-i+t}^j)\Big)}_2^2\notag\\
&\leq \frac{K^2 \sigma^2}{PB}+K\sum_{t=0}^{K-1}\EE\norm{\nabla F(\bw_{n-i+t}^j)}_2^2\notag\\
&\leq  \frac{K^2 \sigma^2}{PB}+K^2M.
\end{align}
Thus, we have the bound of $\EE \norm{v_n}_2^2.$
\begin{align*}
\EE \norm{v_n}_2^2&\leq\Gamma_{n-1}{\sum_{i=0}^{n-1}{{\mu^i} \eta_{n-i} ^2\EE\norm{G_{n-i}}_2^2}}\\&\leq \Gamma_{n-1}{\sum_{i=0}^{n-1}{{\mu^i} \eta_{n-i} ^2\left(\frac{K^2 \sigma^2}{PB}+K^2M\right)}}\\&\leq \frac{\eta^2}{(1-\mu)^2}\left(\frac{K^2 \sigma^2}{PB}+K^2M\right).
\end{align*}

Then we have the bound of the first term.
\begin{align}\label{term1bound}
&\frac{1}{2L}\norm{\nabla F(z_n)-\nabla F(w_n)}_2^2\notag\\\leq& \frac{L}{2}\frac{\mu^2}{(1-\mu)^2}\frac{\eta^2}{(1-\mu)^2}\left(\frac{K^2 \sigma^2}{PB}+K^2M\right).
\end{align}
For the second on the right hand side of \eqref{bound1},  the proof follows that of  Theorem 3.1 in \cite{zhou2017convergence}. We first expend $G_n$ in the second term as follows. 
\begin{align*}
&-\frac{\eta_n}{1-\mu} \nabla F(\tilde{\bw}_n)^\top G_n\\
=&-\frac{\eta_n}{1-\mu}\nabla F(\tilde{\bw}_n)^\top \left(\frac{1}{B_n P}\sum_{j=1}^P\sum_{t=0}^{K-1}\sum_{s=1}^{B_n} \nabla F(\bw_{n+t}^j,\xi_{t,s}^j)\right)\\
=&-\frac{\eta_n}{1-\mu}\frac{1}{B_n P}\sum_{j=1}^P\sum_{t=0}^{K-1}\sum_{s=1}^{B_n} \nabla F(\tilde{\bw}_n)^\top \nabla F(\bw_{n+t}^j,\xi_{t,s}^j).
\end{align*}
Given $\tilde \bw_n,$ for fixed $j$ and $t,$ by the tower property of conditional expectation, we have
\begin{align*}
&\frac{1}{B_n}\sum_{s=1}^{B_n} \EE\nabla F(\tilde{w}_n)^\top \nabla F(\bw_{n+t}^j,\xi_{t,s}^j)\\=&\frac{1}{B_n}\sum_{s=1}^{B_n}\EE\left[\EE\left[ F(\tilde{w}_n)^\top \nabla F(\bw_{n+t}^j,\xi_{t,s}^j)|\bw_{n+t}^j\right]\right]\\
=&\EE\left[F(\tilde{w}_n)^\top \nabla F(\bw_{n+t}^j)\right].
\end{align*} 
Moreover, given $\tilde \bw_n,$ since noise $\xi's$ are i.i.d, and for each $j,$ $\bw_{n+t}^j$ follows the same updated rule, which implies  for any fixed t, $ \nabla F(\bw_{n+t}^j)'s$ are i.i.d.  Thus, we get rid of the summation over $j$ as follows.
\begin{align*}
&\frac{1}{P}\sum_{j=1}^P\frac{1}{B_n}\sum_{s=1}^{B_n} \EE\nabla F(\tilde{w}_n)^\top \nabla F(\bw_{n+t}^j,\xi_{t,s}^j)\\=&\frac{1}{P}\sum_{j=1}^P\EE\left[F(\tilde{w}_n)^\top \nabla F(\bw_{n+t}^j)\right]=\EE\left[F(\tilde{w}_n)^\top \nabla F(\bw_{n+t}^j)\right].
\end{align*}
Together with bound  (3.14)  in \cite{zhou2017convergence}, we get the bound of the expectation of the second term.
\begin{align}\label{term2bound}
&\EE\left[-\frac{\eta_n}{1-\mu}\nabla F(\tilde{\bw}_n)^\top G_n\right]\notag\\\leq&-\frac{(K+1)\eta_n}{2(1-\mu)}\Big(1-\frac{L^2\eta_n^2K(K-1)}{2(1-\mu)^2(K+1)}  \Big) \big\| \nabla F(\widetilde{\bw}_{n})\big\|_2^2\notag\\ &  -\frac{\eta_n}{2(1-\mu)}\Big( 1- \frac{L^2\eta_n^2(K+1)(K-2)}{2(1-\mu^2)}\Big)\sum\limits_{t=1}^{K-1}\EE\Big\|\nabla F(\bw_{n+t}^j)\Big\|_2^2\notag\\&+\frac{L^2\eta_n^3\sigma^2(2K-1)K(K-1)}{12B(1-\mu)^3}.
\end{align}
Following similar lines to \eqref{normbound}, we get the bound of the expectation of the last term.
\begin{align}\label{term3bound}
&\frac{L\eta_n^2}{(1-\mu)^2}\EE \norm{G_n}_2^2\notag\\
\leq & \frac{LK^2\sigma^2 \eta_n^2}{PB(1-\mu)^2}+\frac{\eta_n^2LK}{(1-\mu)^2}\sum\limits_{t=0}^{K-1}\EE\Big\|\nabla F(\bw_{n+t}^j)\Big\|_2^2\notag\\
\leq & \frac{LK^2\sigma^2 \eta_n^2}{PB(1-\mu)^2}+\frac{\eta_n^2LK}{(1-\mu)^2}\sum\limits_{t=1}^{K-1}\EE\Big\|\nabla F(\bw_{n+t}^j)\Big\|_2^2\notag\\&+\frac{\eta_n^2LK}{(1-\mu)^2}\| \nabla F(\widetilde{\bw}_{n})\big\|_2^2.
\end{align}
Combine \eqref{term2bound}, \eqref{term3bound} and \eqref{term1bound} together, and we have
\begin{align*}
&\EE\{F(z_{n+1})-F(z_n)\}\notag\\\leq&
-\frac{(K+1)\eta}{2(1-\mu)}C_1 \EE\big\| \nabla F(\widetilde{\bw}_{n})\big\|_2^2\notag\\ &-\frac{\eta}{2(1-\mu)}C_2\sum\limits_{t=1}^{K-1}\EE\Big\|\nabla F(\bw_{n+t}^j)\Big\|_2^2\notag\\&+\frac{L^2\eta^3\sigma^2(2K-1)K(K-1)}{12B(1-\mu)^3}\notag\\&+\frac{LK^2\sigma^2 \eta^2}{PB(1-\mu)^2}\left(1+\frac{\mu^2}{2(1-\mu)^2}\right)+\frac{L\eta^2\mu^2K^2M}{2(1-\mu)^4},\notag\\ 
\end{align*}
where $C_1=1-\frac{L^2\eta^2K(K-1)}{2(1-\mu)^2(K+1)}-\frac{2\eta LK}{(1-\mu)(K+1)},$ $C_2= 1- \frac{L^2\eta^2(K+1)(K-2)}{2(1-\mu^2)}-\frac{2\eta LK}{1-\mu}.$
Under the condition $1\geq \frac{L^2\eta^2(K+1)(K-2)}{2(1-\mu)^2}-\frac{2\eta LK}{1-\mu},$ i.e., $C_1\geq 0,$ the second term on the right hand side can be discarded. At the same time, the above condition also yields the following inequality.
$$\frac{L^2\eta^2K(K-1)}{2(1-\mu)^2(K+1)}+\frac{2\eta LK}{(1-\mu)(K+1)}\geq \frac{K- L^2\eta^2/(1-\mu)^2}{K+1}.$$
For some $\delta\in(0,1)$ such that $1-\delta\geq L^2\eta^2/(1-\mu)^2,$ we have 
\begin{align*}
&\EE\{F(z_{n+1}) F(z_n)\}\notag\\\leq &-\frac{(K-1+\delta)\eta}{2(1-\mu)} \EE\big\| \nabla F(\widetilde{\bw}_{n})\big\|_2^2+\frac{L^2\eta^3\sigma^2(2K-1)K(K-1)}{12B(1-\mu)^3}\notag\\&+\frac{LK^2\sigma^2 \eta^2}{PB(1-\mu)^2}\left(1+\frac{\mu^2}{2(1-\mu)^2}\right)+\frac{L\eta^2\mu^2K^2M}{2(1-\mu)^4}.
\end{align*}
The above inequality holds for all $i=1,1,...,N$ sum them together and we get
\begin{align*}
&\EE\{F(z_{N+1})-F(z_1)\}\notag\\\leq &-\frac{(K-1+\delta)\eta}{2(1-\mu)} \sum_{i=1}^{N}\EE\big\| \nabla F(\widetilde{\bw}_{i})\big\|_2^2\notag\\&+\frac{NL^2\eta^3\sigma^2(2K-1)K(K-1)}{12B(1-\mu)^3}\notag\\&+\frac{NLK^2\sigma^2 \eta^2}{PB(1-\mu)^2}\left(1+\frac{\mu^2}{2(1-\mu)^2}\right)+\frac{NL\eta^2\mu^2K^2M}{2(1-\mu)^4},
\end{align*}
which is equivalent to 
\begin{align*}
&\frac{(K-1+\delta)\eta}{2(1-\mu)} \sum_{i=1}^{N}\EE\big\| \nabla F(\widetilde{\bw}_{i})\big\|_2^2\notag\\\leq& \EE\{F(z_{1})-F(z_{N+1})\}+\frac{NL^2\eta^3\sigma^2(2K-1)K(K-1)}{12B(1-\mu)^3}\notag\\&+\frac{NLK^2\sigma^2 \eta^2}{PB(1-\mu)^2}\left(1+\frac{\mu^2}{2(1-\mu)^2}\right)+\frac{NL\eta^2\mu^2K^2M}{2(1-\mu)^4}\\
\leq&\EE\{F(z_{1})-F^*\}+\frac{NL^2\eta^3\sigma^2(2K-1)K(K-1)}{12B(1-\mu)^3}\notag\\&+\frac{NLK^2\sigma^2 \eta^2}{PB(1-\mu)^2}\left(1+\frac{\mu^2}{2(1-\mu)^2}\right)+\frac{NL\eta^2\mu^2K^2M}{2(1-\mu)^4}.
\end{align*}
Note that $z_1=w_1$ implies $\EE\{F(z_{1})-F^*\}=F(w_{1})-F^* .$ Multiply each hand side   by $\frac{2(1-\mu)}{N(K-1+\delta)\eta},$ we have
 \begin{align*}
\frac{1}{N} \sum_{i=1}^{N}\EE\big\| \nabla& F(\widetilde{\bw}_{i})\big\|_2^2\leq \frac{2(1-\mu)(F(w_{1})-F^*)}{N(K-1+\delta)\eta}\notag\\&+\frac{L^2\eta^2\sigma^2(2K-1)K(K-1)}{6(K-1+\delta)B(1-\mu)^2}\notag\\
&+\frac{2LK^2\sigma^2 \eta}{PB(K-1+\delta)(1-\mu)}\left(1+\frac{\mu^2}{2(1-\mu)^2}\right)\notag\\&+\frac{L\eta\mu^2K^2M}{(K-1+\delta)(1-\mu)^3}.
\end{align*}
\end{proof}
We have the following observations based on the upper bound \ref{ineq_thm}.
\begin{itemize}
	\item Convergence to $\epsilon-$optimal solution: Note that the first term on the right hand side of \eqref{ineq_thm} goes to  zero as $N\rightarrow\infty,$ while the other term remains unchanged during updating. If we let $\eta\rightarrow 0,$ these terms will vanish. Thus, M-AVG can achieve $\epsilon-$optimal solution where $\epsilon$ can be any positive number given small enough $\eta$ and large $N.$ 
	\item Momentum accelerates convergence but hurts accuracy: The first term is reduced by a factor $1-\mu$ which means M-AVG converges faster than K-AVG. However, adding momentum shows an adverse impact on the final convergence, since the other terms are increased. This observation is constant to that in \cite{liu2018toward}.
	\item Different from K-AVG, the bound of M-AVG has an additional term $ \frac{L\eta\mu^2K^2M}{(K-1+\delta)(1-\mu)^3}$ which is not affected by $P$ and $B$. This is the additional variance induced by the momentum and can only be controlled by the step size. Thus, adding momentum usually requires a small step size. to guarantee the convergence.
\end{itemize}
\subsection{Comparison with K-AVG}

\noindent $\bullet$ Speed Up: Theorem \ref{thm} demonstrates that M-AVG can converge to the $\epsilon-$optimal solution and accelerate the convergence speed. However, as a con of momentum, the final accuracy is increased. A simple question is under the same hyper-parameter tuples $(N,K,P,B),$ does adding momentum leads to a better estimate? In other words, given a target accuracy, does M-SGD achieve it faster than K-AVG? Lemma \ref{lem_opt_mu} provides a positive answer.
It shows that under certain conditions,   the optimal momentum parameter $\mu$ to minimize the upper bound obtained in Theorem \ref{thm} is non-zero .
\begin{lemma}\label{lem_opt_mu}
Under Assumption \ref{assumption}, suppose Algorithm is run with fixed step size $\eta>0,$ batch size $B>0,$ number of learners $P>0a$ for $N$ meta iterations, such that
$$
 1> \frac{L^2\eta^2(K+1)(K-2)}{2}+{2\eta LK}$$
 and\
 $$
 1-\delta> L^2\eta^2,  $$
 for some constant $\delta\in(0,1).$ When the following conditions hold, 
 $$\eta^2< \frac{B(F(w_{1})-F^*)}{5LN\sigma^2(5/P+6L)}~\text{and}~ K\leq 5$$
or $$1>  \frac{N\sigma^2}{2B(F(w_{1})-F^*)}(\frac{1}{2LP}+\frac{1}{L})~\text{and}~ K> 5,$$
 we have  $$\mu_\text{optimal}>0.$$
\end{lemma}
\begin{proof}
	 %Please refer to Appendix \ref{proof_opt_mu}
  Details are omitted to conserve space.
\end{proof}

Lemma \ref{lem_opt_mu} justify that adding momentum does improve the performance of K-AVG under same hyper parameters. The next lemma  show that  M-AVG running $N$ steps can achieve better upper bound than K-AVG running $1/(1-\mu/2)$ steps. For notational simplicity, let the upper bound in Theorem \ref{thm} be $ g(\mu,N,\eta;P,B,K),$ i.e.,
  \begin{align*}
   &g(\mu,N,\eta;P,B,K)\\=& \frac{2(1-\mu)(F(w_{1})-F^*)}{N(K-1+\delta)\eta}\notag\\&+\frac{L^2\eta^2\sigma^2(2K-1)K(K-1)}{6(K-1+\delta)B(1-\mu)^2}\notag\\
  &+\frac{2LK^2\sigma^2 \eta}{PB(K-1+\delta)(1-\mu)}\left(1+\frac{\mu^2}{2(1-\mu)^2}\right)\notag\\&+\frac{L\eta\mu^2K^2M}{(K-1+\delta)(1-\mu)^3}.
  \end{align*}

\begin{lemma}\label{lem_more_N}
Suppose M-AVG is  run with fixed $\mu>0,$ $N,$ $\eta,$ $P,$ $B,$ $K$ such that  $$1\geq \frac{L^2\eta^2(K+1)(K-2)}{2(1-\mu)^2}+\frac{2\eta LK}{1-\mu}$$
and 
$$1-\delta\geq L^2\eta^2/(1-\mu)^2,$$
for some constant $\delta\in(0,1).$ Then under the following condition,
$$\eta<\sqrt{ \frac{PB(F(w_{1})-F^*)(1-\mu)^3}{2NKC_3}},$$
where $C_3=2{LK\sigma^2 }+{PL^2\sigma^2(2K-1)(K-1)}+{L KMPB},$
we have $$ g(\mu,N,\eta;P,B,K)<g(0,\frac{1}{1-\frac{\mu}{2}}N,\eta;P,B,K).$$
\end{lemma}
\begin{proof}
	%Please refer to Appendix \ref{proof_more_N}.
  Details are omitted to conserve space.
\end{proof}

We remark that here $1-\mu/2$ can be replaced by any $1-\alpha \mu$ where $\alpha\in (0,1).$ When $\alpha$ is large, a smaller $\eta$ is required. Lemma \ref{lem_more_N} explicitly show the advantage of M-AVG over K-AVG in terms of convergence speed. This $1-\alpha\mu$ improvement is consistent with the finding in \cite{liu2018toward}. Moreover, this improvement can be seen from the update \eqref{z_update} of the auxiliary sequence. In fact, $z$ can be viewed as the K-AVG update using a larger step size, which leads to a $(1-\alpha \mu)$ faster convergence speed according to Theorem 3.1 in \cite{zhou2017convergence}.  

\noindent $\bullet$ Optimal $K$ is not 1: One important advantage of K-AVG is that frequent averaging is not necessary and the optimal $K$ is proved to be large than 1, which greatly lower the communication cost. We next show that for many cases, this advantage will be inherited even after adding momentum.  We consider the same case as in \cite{zhou2017convergence} where the amount of samples processed $N \times K$ is constant, which means that the computational time remains as a constant given a fixed number of processors. We then have the following lemma.

	\begin{lemma}\label{opt_k_positive}
		Let $S=N*K,$ be a constant. Suppose the Algorithm 1 is run with a fixed step size $\eta,$ a fixed batch size $B,$ and a fixed number of processors $P.$ Suppose for a fixed $\mu\geq0,$ the optimal frequency is $K_{opt}(\mu).$ Then under the following condition,
		\begin{align}\label{conditionk}
		&\frac{1-\delta}{\delta}\frac{(F(w_{1})-F^*)}{S\eta}>\frac{1}{(1-\mu)^3}\frac{L^2\eta^2\sigma^2}{2B}\notag\\&~~+\frac{1}{(1-\mu)^2}\frac{3\delta-1}{2\delta}\left(\frac{\mu^2}{(1-\mu)^2}(\frac{L\sigma^2 \eta}{PB}+{L\eta M})+\frac{2L\sigma^2 \eta}{PB}\right), 
		\end{align}
		we have  $$K_{opt}(\mu)>1.$$
	\end{lemma}
\begin{proof}
	%Please refer to Appendix \ref{proof_k}.
        Details are omitted to conserve space.
		\end{proof}
Our experiment further shows that the optimal $K$ can be  very large such as 32, 64. Thus, M-AVG with large $K$ also enjoys the low communication cost property. Note that condition \eqref{conditionk} requires $(F(w_{1})-F^*)$ to be larger than in Theorem 3.4 in \cite{zhou2017convergence}. Intuitively,  Less averaging and adding momentum both increase the variance, which is only preferred far away from the global solution. As a result, M-AVG need the initialization to be further than K-AVG. 
\subsection{Tuning guidelines}
We have shown M-AVG can achieve great performance. However, since M-AVG has 5 hyper parameters $(P,B,K,\eta,\mu)$ which need tuning, it is still difficult to achieve the best performance of M-AVG in practice without any guidelines. In this section we provide two tuning guidelines theoretically to ease the work or practitioners.

\noindent $\bullet$ More processors available (Increase P):  A nature question is if we have more processors, how to change $\mu$ accordingly. Let the original number of processors be $P_0.$ We consider the case that the total number of samples processed $N*P*B*K$ is a constant and $B,K$ are fixed. We remark that, in the following discussion, we assume the condition in Theorem \ref{thm} holds, i.e.,
  $$1\geq \frac{L^2\eta^2(K+1)(K-2)}{2(1-\mu)^2}-\frac{2\eta LK}{1-\mu}$$and $$1-\delta\geq L^2\eta^2/(1-\mu)^2,$$ hold for some $\delta\in(0,1).$ We then have the following lemma.
 \begin{lemma}\label{lem_increase_P}
 Let $S=N*P*B*K,$ be a constant. Suppose the Algorithm 1 is run with a fixed step size $\eta,$ a fixed batch size $B,$ and a fixed frequency $K.$ Suppose for  $P=P_0,$  the optimal momentum parameter is $\mu_0^*.$ If the number of processors is increased from $P_0$ to $\lambda P_0,$ where $\lambda>1,$ the momentum parameter $\mu_{\lambda}^*$ satisfies
 $$\mu_{\lambda}^*>\mu_0^*.$$
 \end{lemma}
 \begin{proof}
 	%Please refer to Appendix \ref{pf_p}.
   Details are omitted to conserve space.
 \end{proof}
Intuitively, increasing number of processors will decrease the noise in the average stochastic gradient. Thus, M-AVG can tolerate more variance from a larger momentum. At the same time, since the number of meta iterations decrease, a larger momentum also helps improve the performance.

\noindent $\bullet$ Switch From K-AVG to M-AVG: We next show that adding momentum can decrease the optimal $K.$ We consider the same setting as in Lemma \ref{opt_k_positive}.
 \begin{lemma}\label{opt_k}
	Let $S=N*K,$ be a constant. Suppose the Algorithm 1 is run with a fixed step size $\eta,$ a fixed batch size $B,$ and a fixed number of processors $P.$ Suppose for a fixed $\mu\geq0,$ the optimal frequency is $K_{opt}(\mu).$ Then under the following condition,
\begin{align}
&\frac{1-\delta}{\delta}\frac{(F(w_{1})-F^*)}{S\eta}>\frac{1}{(1-\mu)^3}\frac{L^2\eta^2\sigma^2}{2B}\notag\\&~~+\frac{1}{(1-\mu)^2}\frac{3\delta-1}{2\delta}\left(\frac{\mu^2}{(1-\mu)^2}(\frac{L\sigma^2 \eta}{PB}+{L\eta M})+\frac{2L\sigma^2 \eta}{PB}\right), 
\end{align}
  we have  $$K_{opt}(\mu)\leq K_{opt}(0).$$
 \end{lemma}
As we discussed after Lemma \ref{opt_k_positive}, frequent averaging and momentum both increase the variance, and thus are conflict. As a direct result, we need to decrease $K$ when $\mu$ is increased.

Both of these two tuning guidelines will be empirically verified in Section \ref{experiment}. There are many other cases that worth consideration. Due to space limit, we only present these two and leave others to our future research.
\section{Experiment}\label{experiment}

We experiment on a cluster of IBM Power9 CPUs with NVIDIA Volta100
GPUs connected with Infiniband.  We use Spectrum-MPI for
communication. All algorithms are implemented with PyTorch~\cite{PGC17}.  We test our implementation with image classification using the \cifar data set and the \imagenet data set. 

\subsection{\mavg accelerates convergence}
\begin{figure}[h]
     \centering
      \includegraphics[width=.50\textwidth]{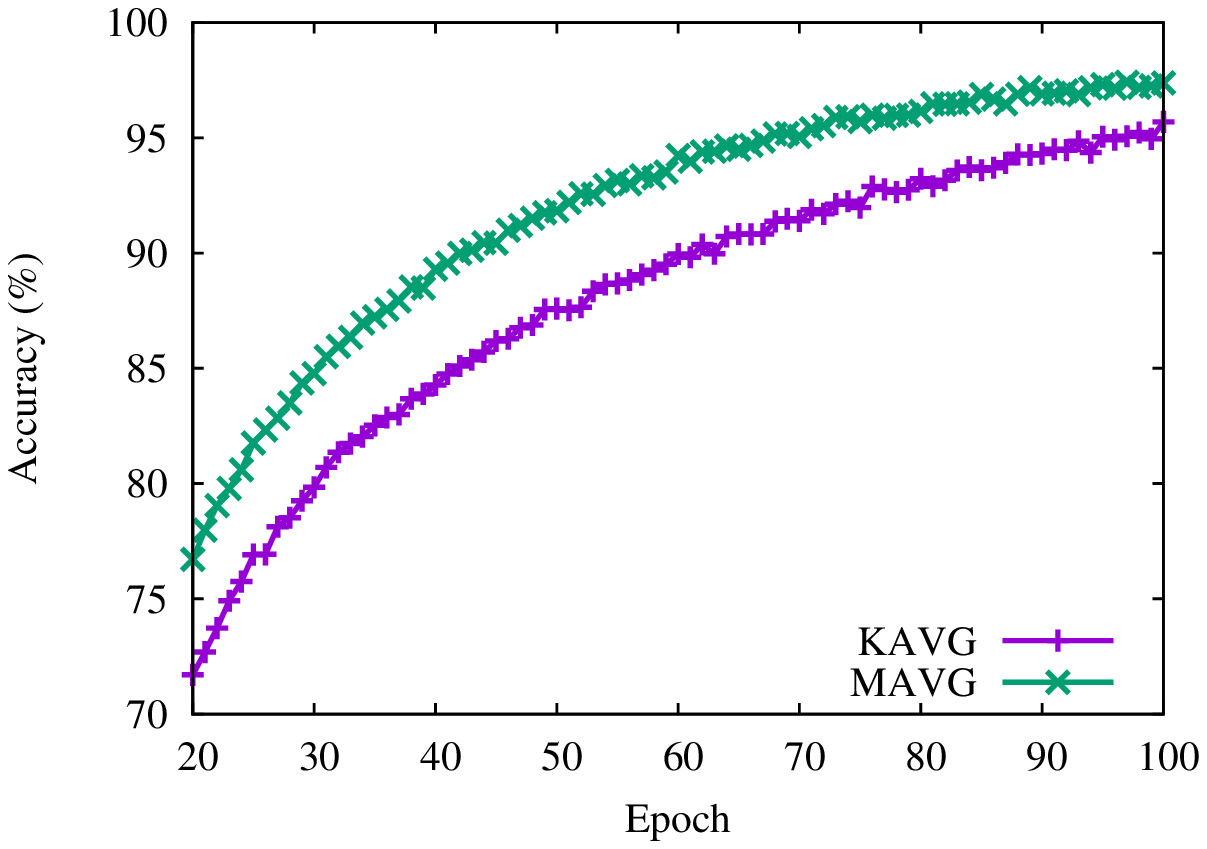}
      \caption{\resnet}
      \label{fig:resnet-train}
  \end{figure}

\begin{figure}[h]
     \centering
      \includegraphics[width=.50\textwidth]{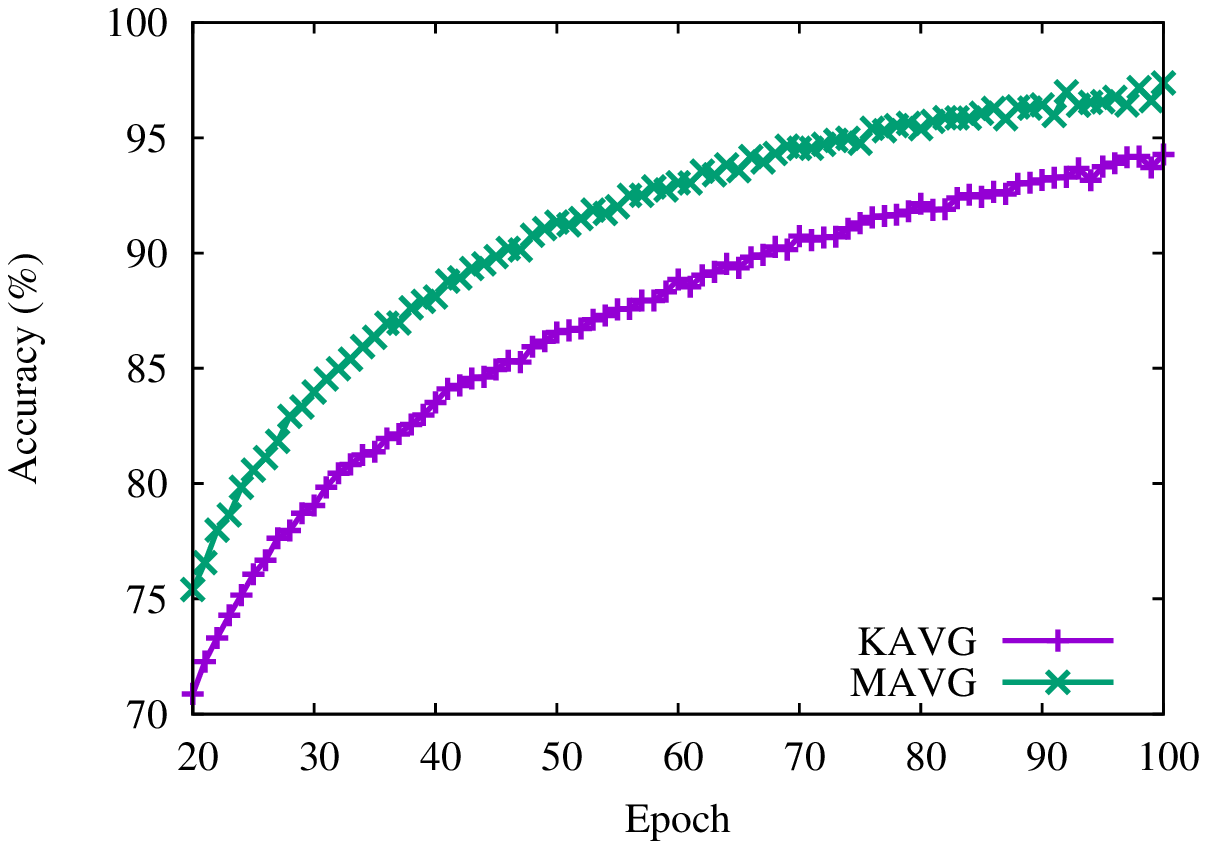}
      \caption{\preact}
      \label{fig:preact-train}
  \end{figure}

\begin{figure}[h]
     \centering
      \includegraphics[width=.50\textwidth]{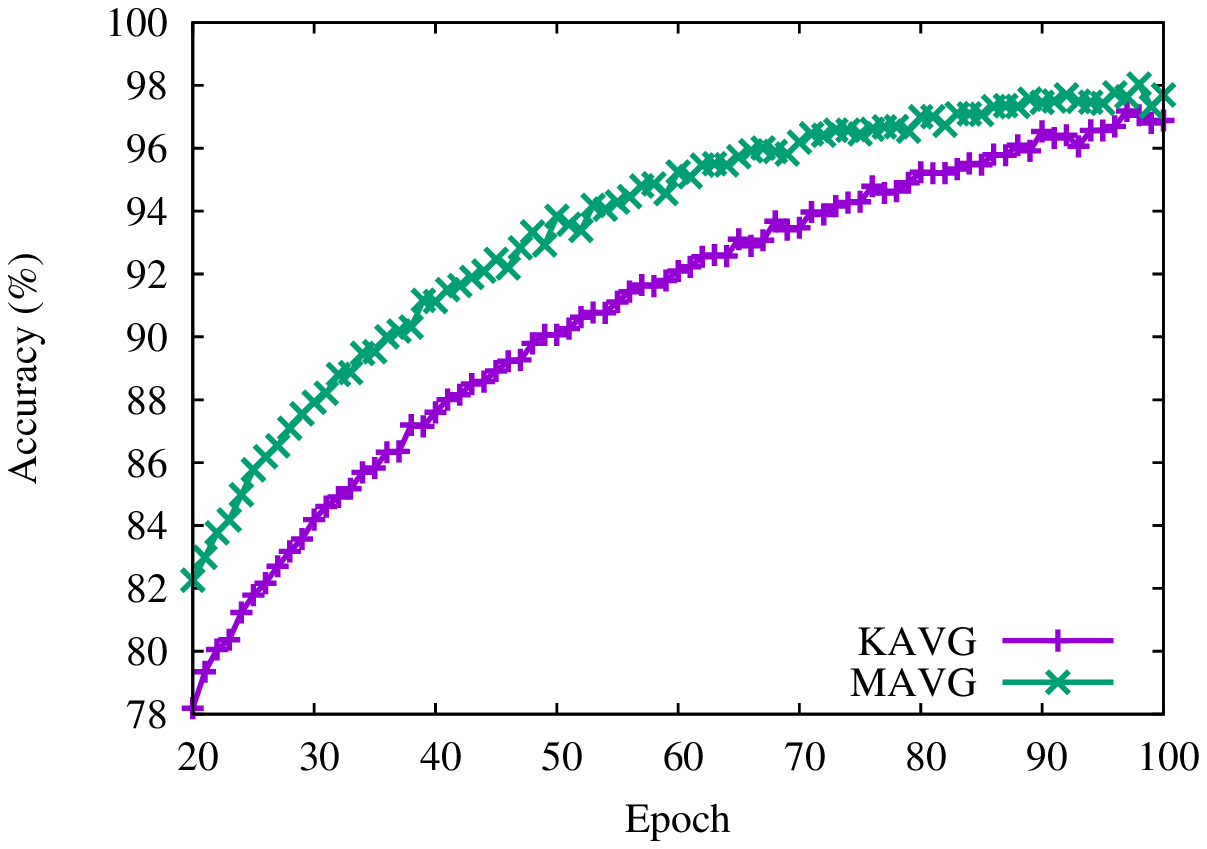}
      \caption{\googlenet}
      \label{fig:googlenet-train}
  \end{figure}

\begin{figure}[h]
     \centering
      \includegraphics[width=.50\textwidth]{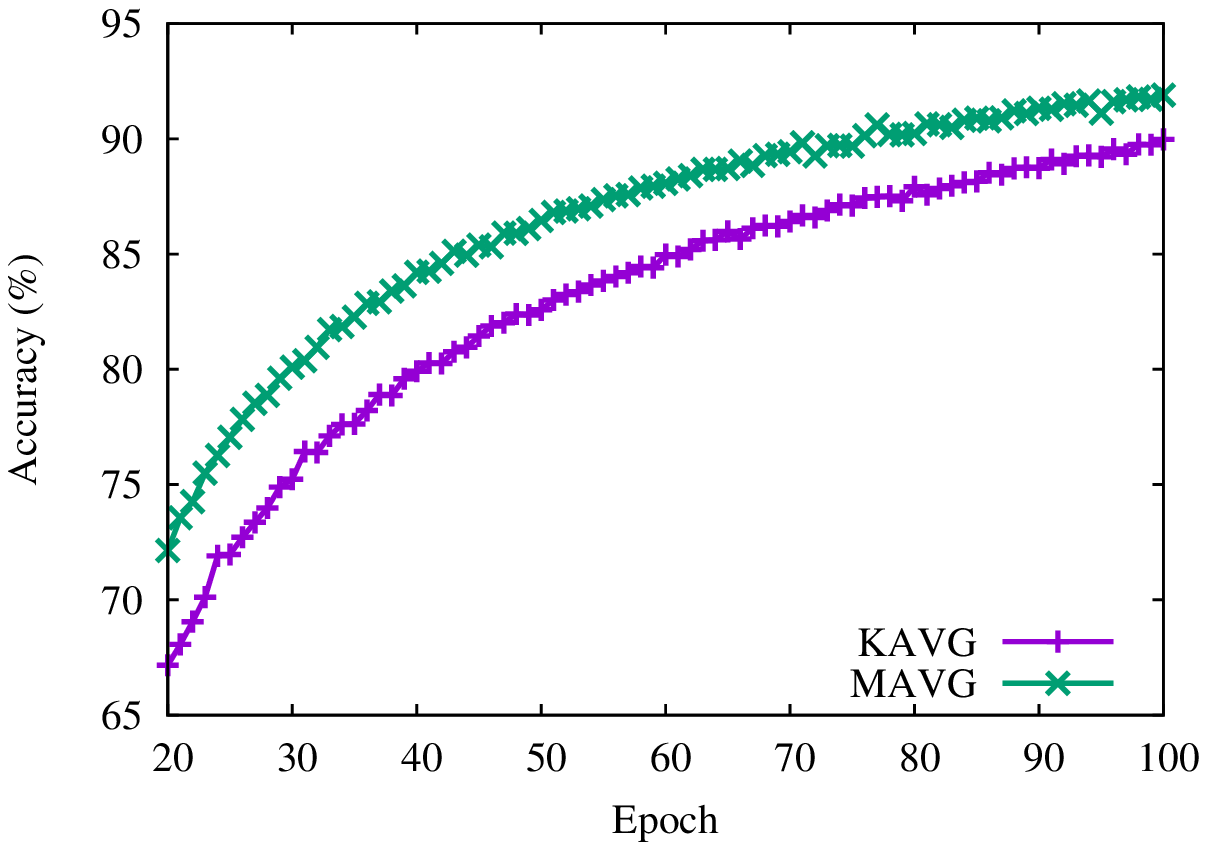}
      \caption{\mobilenet}
      \label{fig:mobilenet-train}
  \end{figure}

\begin{figure}[h]
     \centering
      \includegraphics[width=.50\textwidth]{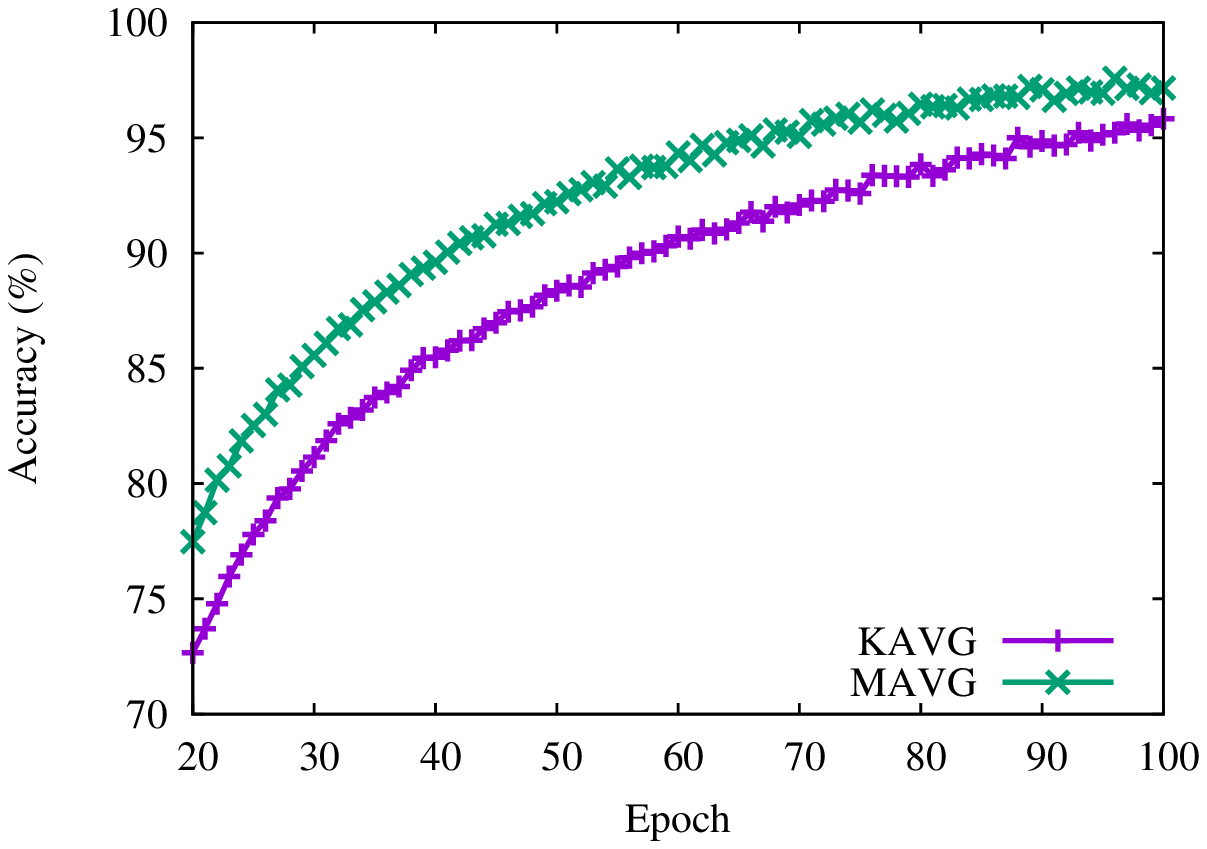}
      \caption{\dpn}
      \label{fig:dpn92-train}
  \end{figure}

\begin{figure}[h]
     \centering
      \includegraphics[width=.50\textwidth]{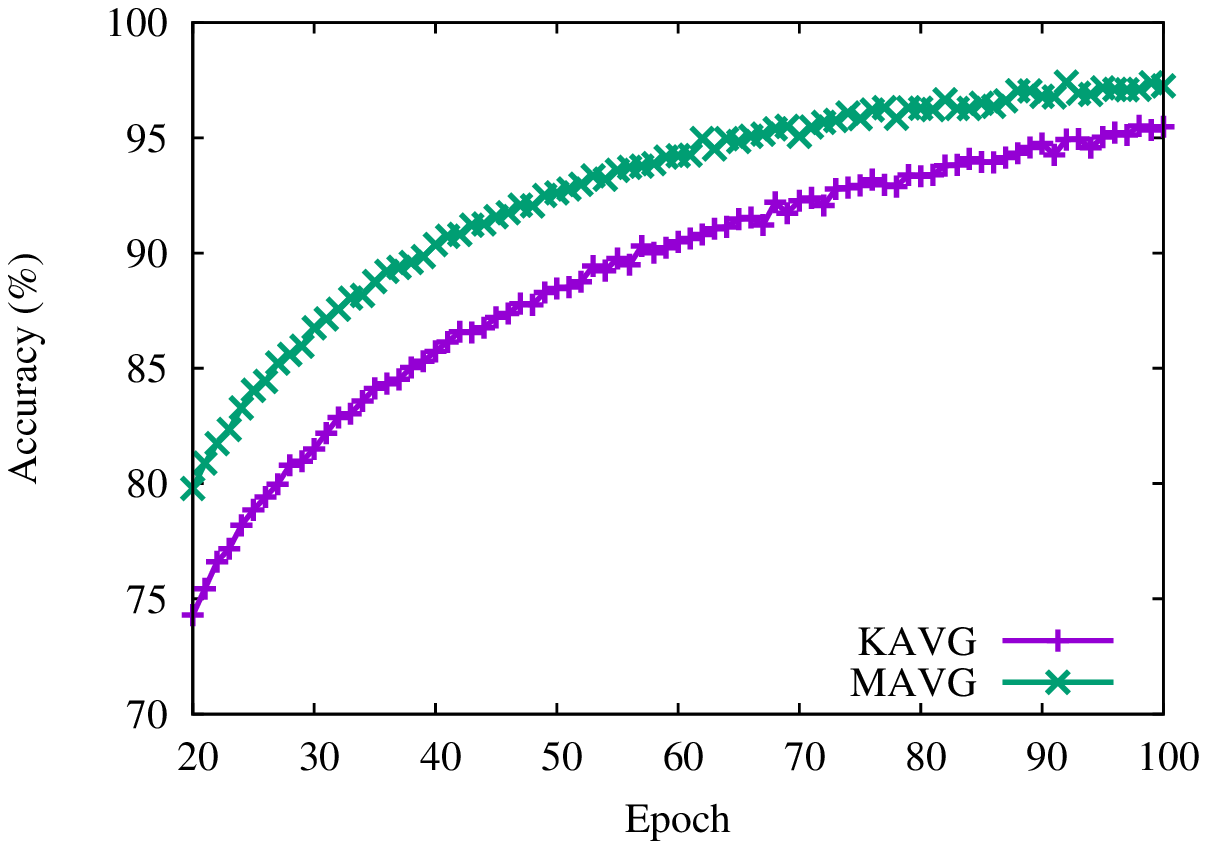}
      \caption{\senet}
      \label{fig:senet-train}
  \end{figure}

We first demonstrate that \mavg accelerates convergence using a range of neural network models with the \cifar data set. 
The 7 state-of-art neural network models we use in our study are: \resnet~\cite{HZR16a}, \densenet~\cite{HLW16}, \senet~\cite{HSS17},
\googlenet~\cite{SLJ14}, \mobilenet~\cite{HZC17},
\preact~\cite{HZR16a}, and \dpn~\cite{CLX17}. They represent some of
the most advanced convolution neural network (CNN) architectures used
in current computer vision tasks. Figures~\ref{fig:resnet-train},~\ref{fig:preact-train},~\ref{fig:googlenet-train},~\ref{fig:mobilenet-train},~\ref{fig:mobilenet-train},~\ref{fig:dpn92-train},~\ref{fig:senet-train} show the evolution of training accuracies for \kavg and \mavg.  

% \begin{table}[htbp]
%  \begin{center}
%  \begin{tabular}{|| c || l| l | l  || }
%  \hline
%  Model& \kavg &\mavg \\ \hline
%  \resnet &  99.654&  99.984 \\ \hline
%  \densenet &       99.992&  99.994 \\ \hline
%  \senet&         99.944&  99.968 \\ \hline
% \googlenet &  99.984&  99.992 \\ \hline
%  \mobilenet &       98.666&  98.966 \\ \hline
%  \preact&         99.966&  99.918 \\ \hline
% \dpn&         99.972&  99.984 \\ \hline
%  \end{tabular}
%  \end{center}
%  \caption{Training accuracy}
%  \label{tab:stall}
% \end{table}
In the figure, the labels \emph{res}, \emph{pre}, \emph{goo},
\emph{den}, \emph{mob}, \emph{dpn}, and \emph{sen} are for \resnet,
\preact, \googlenet, \densenet, \mobilenet, \dpn, and \senet,
respectively. 

Table~\ref{tab:val-cifar} shows the validation accuracy achieved after 200 epochs. 

\begin{table}[htbp]
 \begin{center}
 \begin{tabular}{|| c || l| l | l  || }
 \hline
 Model& \kavg &\mavg \\ \hline
 \resnet &  94.81&  95.31 \\ \hline
 \densenet &      95.2&  95.5 \\ \hline
 \senet&        94.73&  94.91 \\ \hline
\googlenet &  94.36&  95.00 \\ \hline
 \mobilenet &       91.77&  92.16 \\ \hline
 \preact&         94.54&  95.03 \\ \hline
\dpn&         95.69&  95.75 \\ \hline
 \end{tabular}
 \end{center}
 \caption{Validation accuracy}
 \label{tab:val-cifar}
\end{table}

\begin{figure}[h]
     \centering
      \includegraphics[width=.50\textwidth]{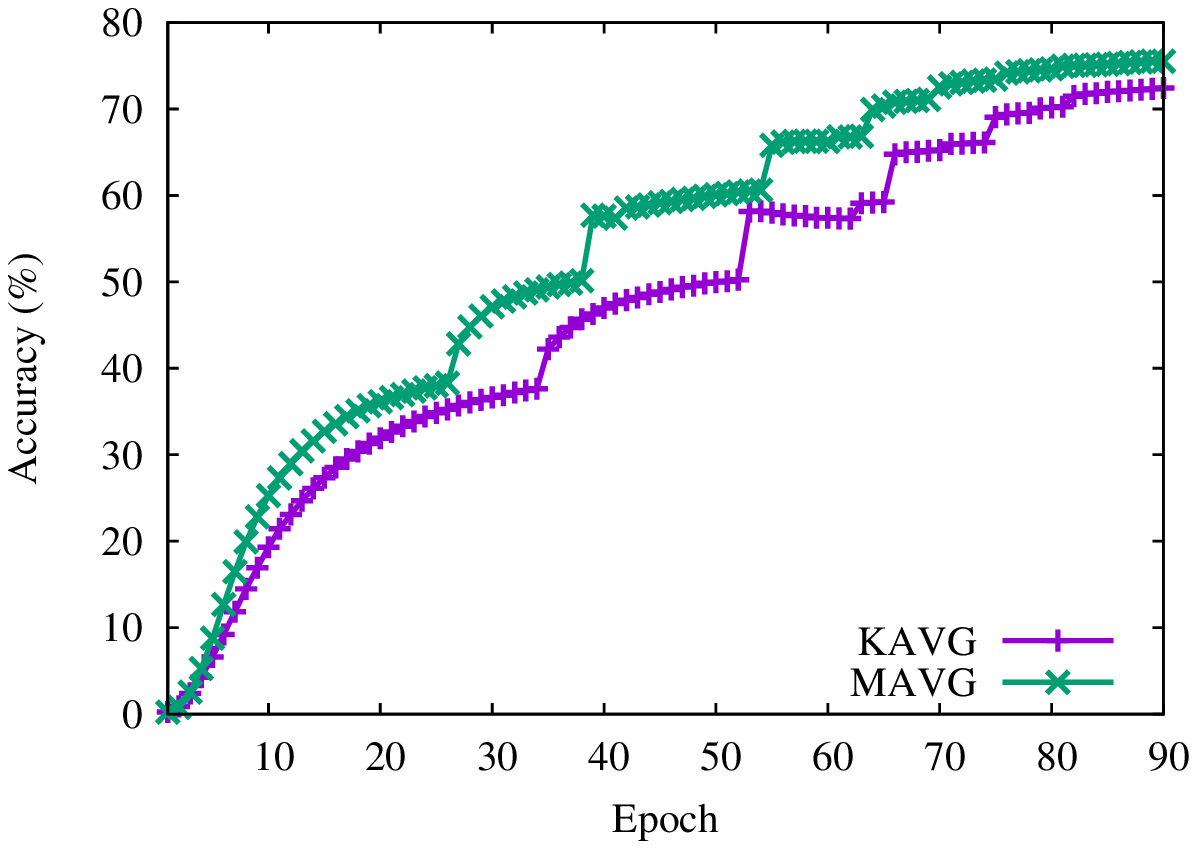}
      \caption{ResNet50}
      \label{fig:imagenet-train}
  \end{figure}

\begin{figure}[h]
     \centering
      \includegraphics[width=.50\textwidth]{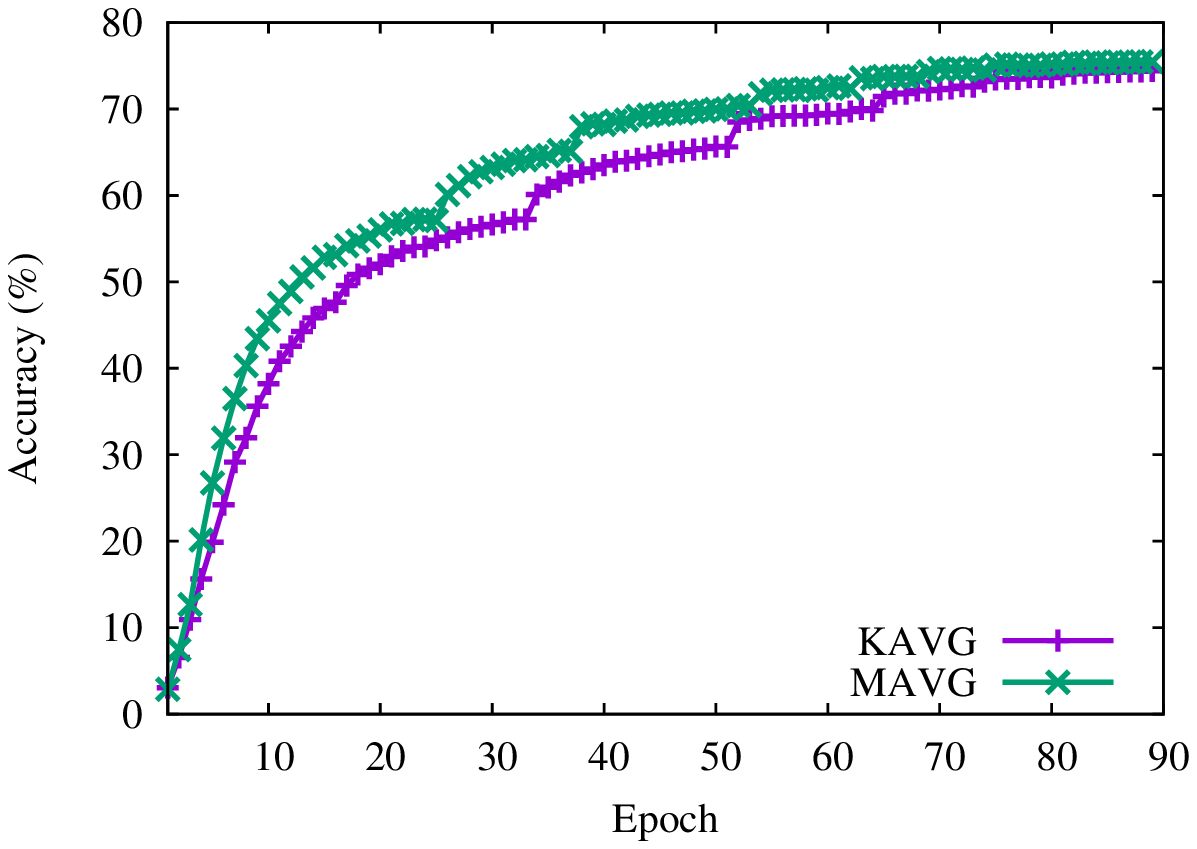}
      \caption{ResNet50}
      \label{fig:imagenet-val}
  \end{figure}

Figures~\ref{fig:imagenet-train} and ~\ref{fig:imagenet-val} shows the training and validation accuracy using ResNet-50 with ImageNet-1K. Again \mavg performs better than \kavg, and this demonstrates the acceleration of convergence through momentum. 

\subsection{$\mu$ with P and an optimal $\mu$}
\begin{figure}[h]
     \centering
      \includegraphics[width=.50\textwidth]{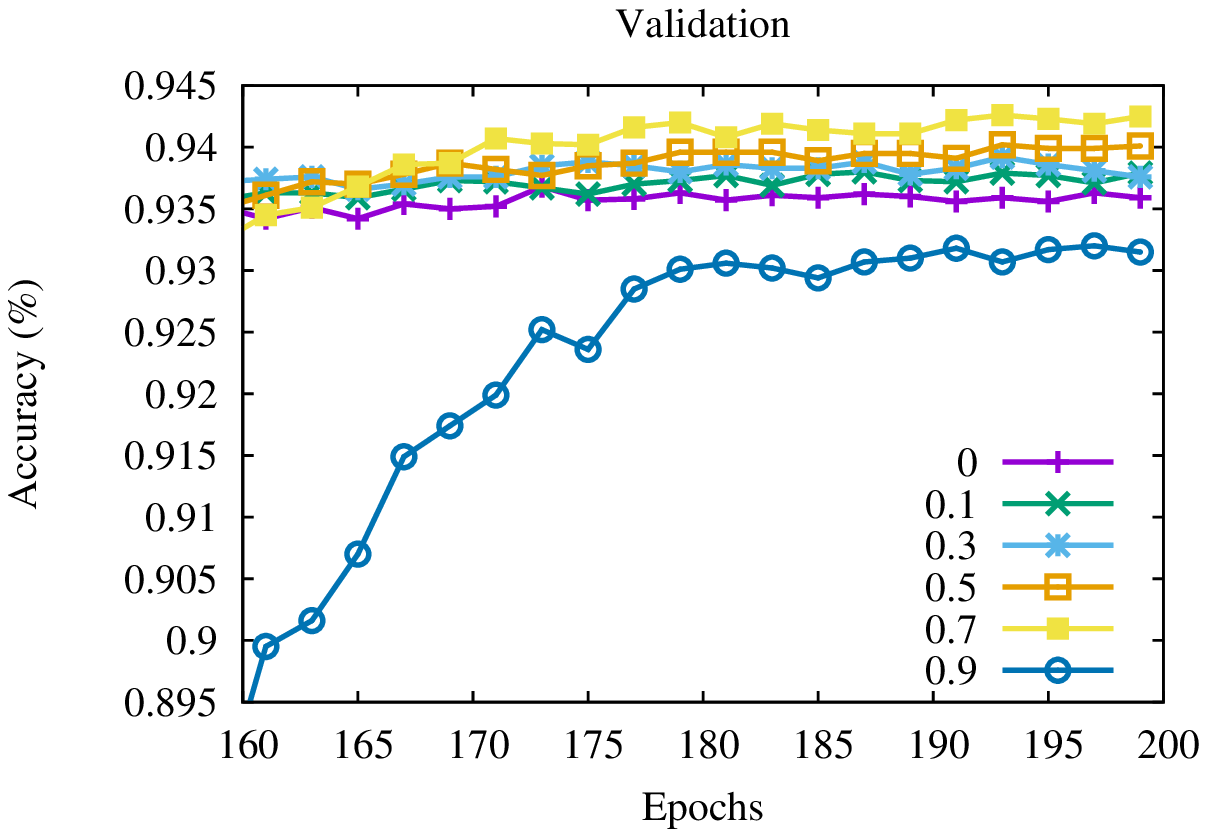}
      \caption{P=6}
      \label{fig:P=6}
  \end{figure}

\begin{figure}[h]
     \centering
      \includegraphics[width=.50\textwidth]{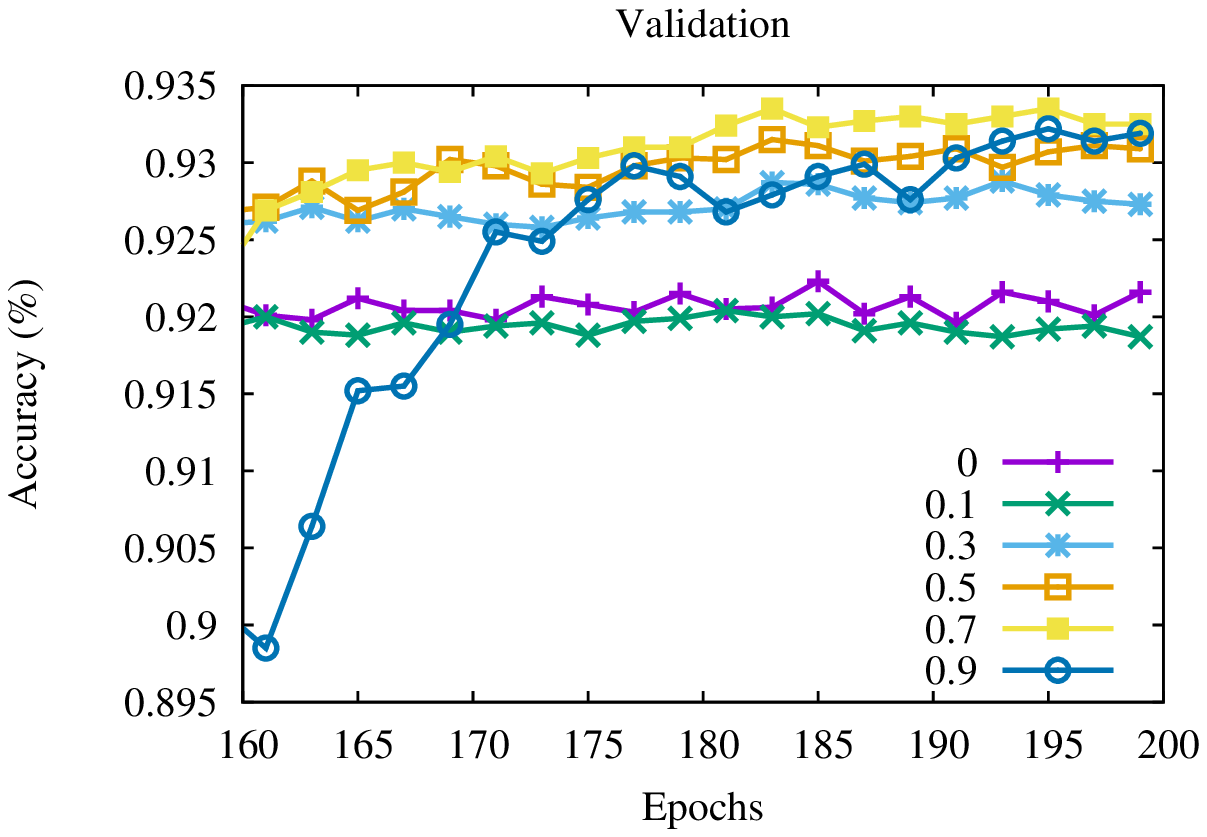}
      \caption{P=12}
      \label{fig:P=12}
  \end{figure}

\begin{figure}[h]
     \centering
      \includegraphics[width=.50\textwidth]{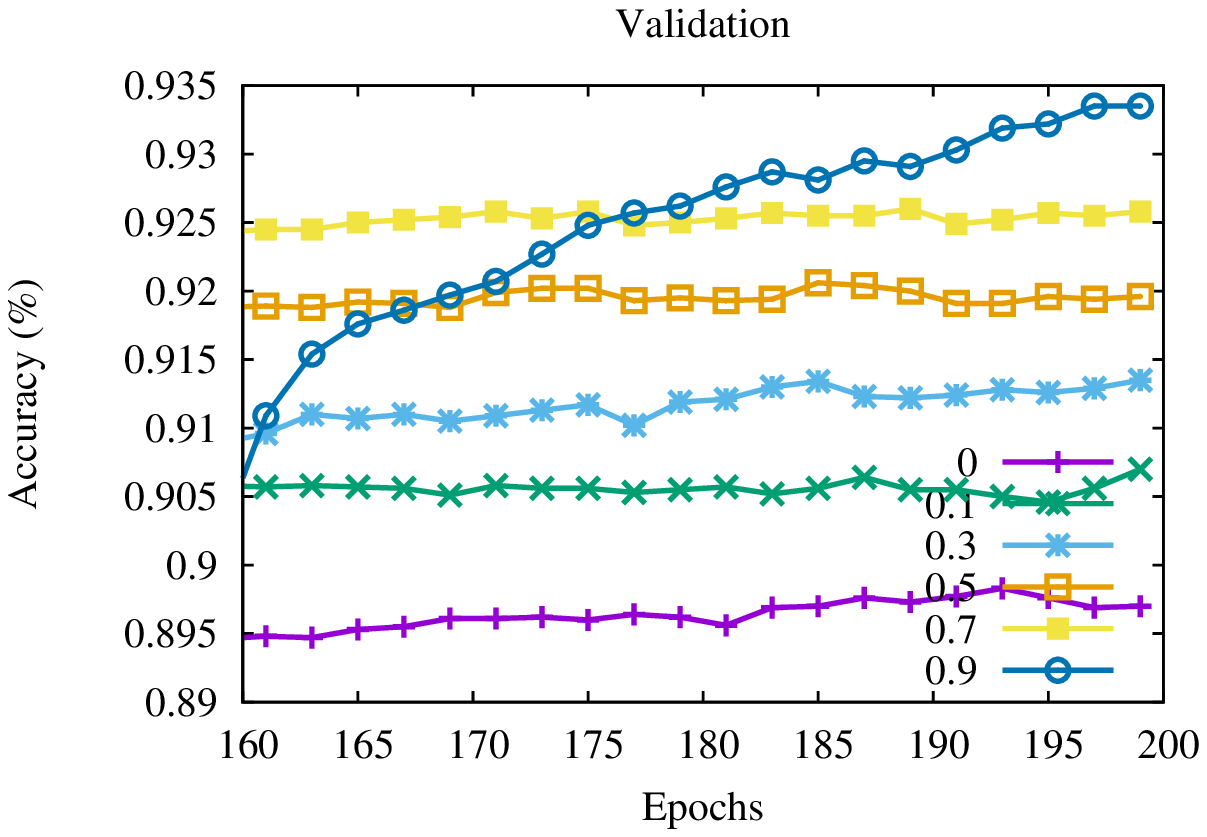}
      \caption{P=24}
      \label{fig:P=24}
  \end{figure}

\begin{figure}[h]
     \centering
      \includegraphics[width=.50\textwidth]{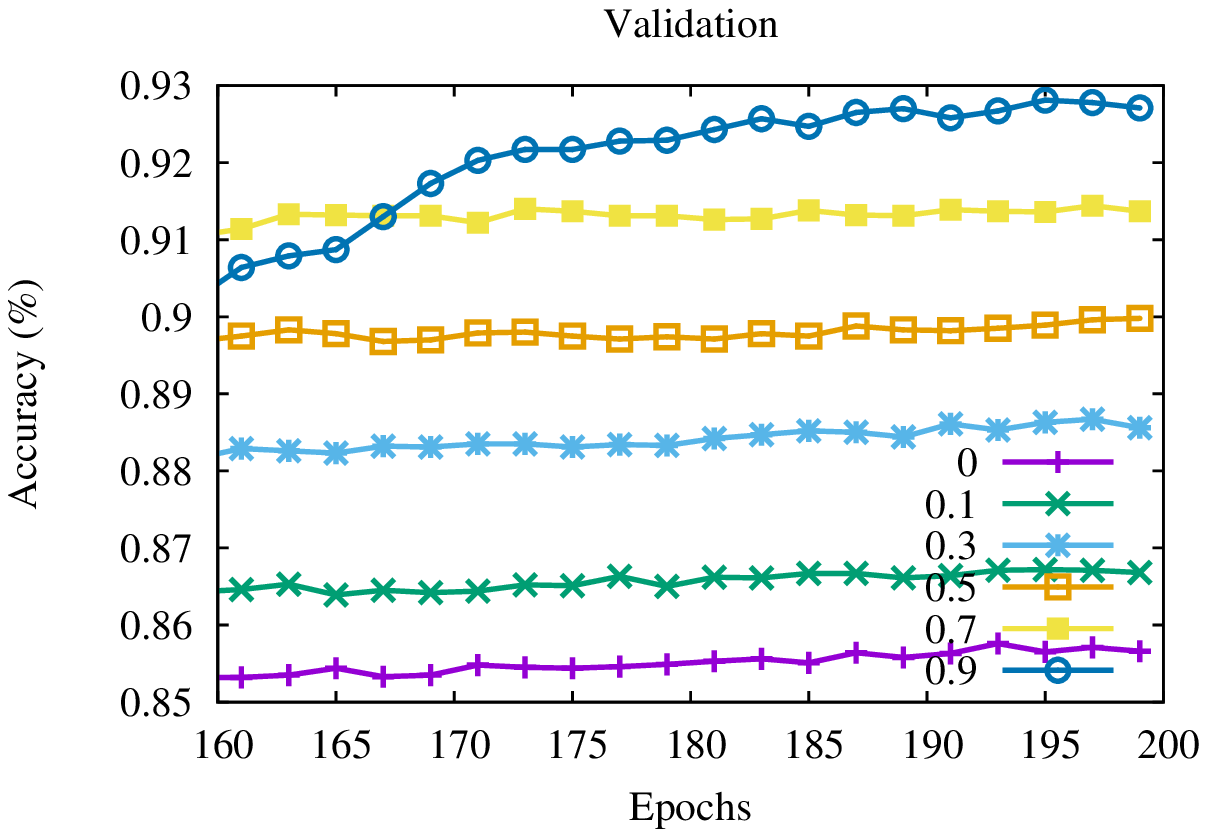}
      \caption{P=48}
      \label{fig:P=48}
  \end{figure}

we study how to tune $\mu$ when we have more processors. We use ResNet18 with the \cifar data set. We set $P=6,12,24,48$ and for each choice of $P$, we test different $\mu$'s ranged from $0$ to $0.9.$ The validation accuracies are shown in Figures~\ref{fig:P=6},~\ref{fig:P=12},~\ref{fig:P=24},~\ref{fig:P=48}. When $P$ is small, we can see the optimal $\mu$ is 0.7. When $\mu<0.7,$ the larger $\mu$ is, the higher accuracy we can achieve. However, $\mu=0.9$  works the worst among all the choices. This is because the additional variance caused by momentum is too large and ruin the performance. However, as $P$ increases, $\mu=0.9$ becomes a better choice  gradually. It is the best choice when $P=48$. These observation is consistent with the tuning guidelines we suggest in Lemma \ref{lem_increase_P} that we can use a lager $\mu$ when we have more processors. 

% \subsection{K}

% \begin{figure}[h]
%      \centering
%       \includegraphics[width=.50\textwidth]{./data1/K-bm.eps}
%       \caption{K}
%       \label{K}
%   \end{figure}

\section{Conclusions}\label{conclusion}
In this paper, we propose a new momentum method M-AVG and theoretically prove the finite sample error bound. Our theory justifies the speed up brought by adding momentum. Thus, M-AVG can achieve faster convergence than K-AVG. At the same time the optimal $K$ is large than 1 which implies M-AVG has low communication cost. We also provide two useful tuning guidelines. Specifically, when we have more processors , we can choose a larger momentum. When we switch from K-AVG to M-AVG, we have use a smaller K.

Our M-AVG algorithm adds momentum at the meta level. We can certainly use MSGD instead of simple SGD in the learner level to accelerate. Our numerical experiment find that this algorithm does have a good performance. However, it is theoretically much harder to analyze its behavior. We leave it to our future research.
\newpage
 \bibliographystyle{plain}
  \bibliography{refer}

\begin{thebibliography}{10}

\bibitem{chen2016scalable}
Kai Chen and Qiang Huo.
\newblock Scalable training of deep learning machines by incremental block
  training with intra-block parallel optimization and blockwise model-update
  filtering.
\newblock In {\em 2016 ieee international conference on acoustics, speech and
  signal processing (icassp)}, pages 5880--5884. IEEE, 2016.

\bibitem{CLX17}
Yunpeng Chen, Jianan Li, Huaxin Xiao, Xiaojie Jin, Shuicheng Yan, and Jiashi
  Feng.
\newblock Dual path networks.
\newblock {\em CoRR}, abs/1707.01629, 2017.

\bibitem{dean2012large}
Jeffrey Dean, Greg Corrado, Rajat Monga, Kai Chen, Matthieu Devin, Mark Mao,
  Andrew Senior, Paul Tucker, Ke~Yang, Quoc~V Le, et~al.
\newblock Large scale distributed deep networks.
\newblock In {\em Advances in neural information processing systems}, pages
  1223--1231, 2012.

\bibitem{ghadimi2016accelerated}
Saeed Ghadimi and Guanghui Lan.
\newblock Accelerated gradient methods for nonconvex nonlinear and stochastic
  programming.
\newblock {\em Mathematical Programming}, 156(1-2):59--99, 2016.

\bibitem{HZR16a}
Kaiming He, Xiangyu Zhang, Shaoqing Ren, and Jian Sun.
\newblock Deep residual learning for image recognition.
\newblock In {\em Proceedings of the IEEE conference on computer vision and
  pattern recognition}, pages 770--778, 2016.

\bibitem{HZC17}
Andrew~G. Howard, Menglong Zhu, Bo~Chen, et~al.
\newblock {MobileNets}: Efficient convolutional neural networks for mobile
  vision applications.
\newblock {\em CoRR}, abs/1704.04861, 2017.

\bibitem{HSS17}
Jie Hu, Li~Shen, and Gang Sun.
\newblock Squeeze-and-excitation networks.
\newblock {\em CoRR}, abs/1709.01507, 2017.

\bibitem{HLW16}
Gao Huang, Zhuang Liu, and Kilian~Q. Weinberger.
\newblock Densely connected convolutional networks.
\newblock {\em CoRR}, abs/1608.06993, 2016.

\bibitem{KP15}
Janis Keuper and Franz{-}Josef Pfreundt.
\newblock Asynchronous parallel stochastic gradient descent - {A} numeric core
  for scalable distributed machine learning algorithms.
\newblock {\em CoRR}, abs/1505.04956, 2015.

\bibitem{lian2015asynchronous}
Xiangru Lian, Yijun Huang, Yuncheng Li, and Ji~Liu.
\newblock Asynchronous parallel stochastic gradient for nonconvex optimization.
\newblock In {\em Advances in Neural Information Processing Systems}, pages
  2737--2745, 2015.

\bibitem{lin2018don}
Tao Lin, Sebastian~U Stich, and Martin Jaggi.
\newblock Don't use large mini-batches, use local sgd.
\newblock {\em arXiv preprint arXiv:1808.07217}, 2018.

\bibitem{liu2018toward}
Tianyi Liu, Zhehui Chen, Enlu Zhou, and Tuo Zhao.
\newblock Toward deeper understanding of nonconvex stochastic optimization with
  momentum using diffusion approximations.
\newblock {\em arXiv preprint arXiv:1802.05155}, 2018.

\bibitem{liu2018towards}
Tianyi Liu, Shiyang Li, Jianping Shi, Enlu Zhou, and Tuo Zhao.
\newblock Towards understanding acceleration tradeoff between momentum and
  asynchrony in nonconvex stochastic optimization.
\newblock In {\em Advances in Neural Information Processing Systems}, pages
  3682--3692, 2018.

\bibitem{ochs2015ipiasco}
Peter Ochs, Thomas Brox, and Thomas Pock.
\newblock ipiasco: Inertial proximal algorithm for strongly convex
  optimization.
\newblock {\em Journal of Mathematical Imaging and Vision}, 53(2):171--181,
  2015.

\bibitem{PGC17}
Adam Paszke, Sam Gross, Soumith Chintala, et~al.
\newblock Automatic differentiation in pytorch.
\newblock 2017.

\bibitem{polyak1964some}
Boris~T Polyak.
\newblock Some methods of speeding up the convergence of iteration methods.
\newblock {\em USSR Computational Mathematics and Mathematical Physics},
  4(5):1--17, 1964.

\bibitem{recht2011hogwild}
Benjamin Recht, Christopher Re, Stephen Wright, and Feng Niu.
\newblock Hogwild: A lock-free approach to parallelizing stochastic gradient
  descent.
\newblock In {\em Advances in neural information processing systems}, pages
  693--701, 2011.

\bibitem{robbins1951stochastic}
Herbert Robbins and Sutton Monro.
\newblock A stochastic approximation method.
\newblock {\em The annals of mathematical statistics}, pages 400--407, 1951.

\bibitem{SLJ14}
Christian Szegedy, Wei Liu, Yangqing Jia, Pierre Sermanet, et~al.
\newblock Going deeper with convolutions.
\newblock {\em CoRR}, abs/1409.4842, 2014.

\bibitem{wilson2016lyapunov}
Ashia~C Wilson, Benjamin Recht, and Michael~I Jordan.
\newblock A lyapunov analysis of momentum methods in optimization.
\newblock {\em arXiv preprint arXiv:1611.02635}, 2016.

\bibitem{yang2016unified}
Tianbao Yang, Qihang Lin, and Zhe Li.
\newblock Unified convergence analysis of stochastic momentum methods for
  convex and non-convex optimization.
\newblock {\em arXiv preprint arXiv:1604.03257}, 2016.

\bibitem{yu2019linear}
Hao Yu, Rong Jin, and Sen Yang.
\newblock On the linear speedup analysis of communication efficient momentum
  sgd for distributed non-convex optimization.
\newblock {\em arXiv preprint arXiv:1905.03817}, 2019.

\bibitem{zhou2017convergence}
Fan Zhou and Guojing Cong.
\newblock On the convergence properties of a $ k $-step averaging stochastic
  gradient descent algorithm for nonconvex optimization.
\newblock {\em arXiv preprint arXiv:1708.01012}, 2017.

\bibitem{zinkevich2010parallelized}
Martin Zinkevich, Markus Weimer, Lihong Li, and Alex~J Smola.
\newblock Parallelized stochastic gradient descent.
\newblock In {\em Advances in neural information processing systems}, pages
  2595--2603, 2010.

\end{thebibliography}

\end{document}